\pdfoutput=1

\documentclass[11pt]{article}

\usepackage{listings}
\usepackage{courier}

\usepackage[final]{acl}

\usepackage{times}
\usepackage{latexsym}

\usepackage[T1]{fontenc}

\usepackage[utf8]{inputenc}

\usepackage{microtype}

\usepackage{inconsolata}

\usepackage{graphicx}

\usepackage{booktabs, multirow}

\usepackage{ulem}

\DeclareRobustCommand{\katoerase}{\bgroup\markoverwith{\textcolor[rgb]{0,0,0.9}{\rule[.5ex]{2pt}{0.8pt}}}\ULon}

\usepackage[most]{tcolorbox}
\usepackage{float}
\usepackage{xspace}
\usepackage{caption}
\tcbset{
  aiboxmain/.style={
    width=\columnwidth,
    top=10pt,
    colback=white,
    colframe=black,
    colbacktitle=black,
    enhanced,
    center,
    attach boxed title to top left={yshift=-0.1in,xshift=0.15in},
    boxed title style={boxrule=0pt,colframe=white,},
  }
}
\newtcolorbox{AIboxmain}[2][]{aiboxmain,title=#2,#1}

\tcbset{
  aibox/.style={
    width=1.9\columnwidth,
    top=10pt,
    colback=white,
    colframe=black,
    colbacktitle=black,
    enhanced,
    center,
    attach boxed title to top left={yshift=-0.1in,xshift=0.15in},
    boxed title style={boxrule=0pt,colframe=white,},
  }
}
\newtcolorbox{AIbox}[2][]{aibox,title=#2,#1}

\title{MK2 at PBIG Competition: A Prompt Generation Solution}

\author{
 \textbf{Yuzheng Xu\textsuperscript{1,2}},
 \textbf{Tosho Hirasawa\textsuperscript{1,2}},
 \textbf{Seiya Kawano\textsuperscript{3,2}},
 \textbf{Shota Kato\textsuperscript{4}},
 \textbf{Tadashi Kozuno\textsuperscript{1,2}}
\\
\\
 \textsuperscript{1}OMRON SINIC X,
 \textsuperscript{2}NexaScience,
 \textsuperscript{3}Kyoto Institute of Technology,
 \textsuperscript{4}Kyoto University
\\
 \small{
   \textbf{Correspondence:} \href{mailto:yuzheng.xu@sinicx.com}{yuzheng.xu@sinicx.com}
 }
}

\begin{document}
\maketitle
\begin{abstract}
The Patent-Based Idea Generation task asks systems to turn real patents into product ideas viable within three years. We propose \textbf{MK2},  a prompt-centric pipeline: Gemini 2.5 drafts and iteratively edits a prompt, grafting useful fragments from weaker outputs; GPT-4.1 then uses this prompt to create one idea per patent, and an Elo loop judged by Qwen3-8B selects the best prompt—all without extra training data. Across three domains, two evaluator types, and six criteria, MK2 topped the automatic leaderboard and won 25 of 36 tests. Only the materials-chemistry track lagged, indicating the need for deeper domain grounding; yet, the results show that lightweight prompt engineering has already delivered competitive, commercially relevant ideation from patents.
\end{abstract}

\section{Introduction}
Large language models (LLMs) have progressed from factual question answering to tasks that demand creativity and domain knowledge. Recent work shows that LLMs can suggest novel scientific hypotheses~\citep{wang-etal-2024-scimon,si2025can}, yet their capacity for commercial ideation remains less understood~\citep{meincke2024using}. The Patent-Based Idea Generation (PBIG) competition~\citep{agentscen2025pbig} addresses this gap by asking systems to transform patent disclosures into market-ready product concepts and by evaluating those concepts with both AI and human judges.

We answer this challenge with a lightweight pipeline that casts product ideation as prompt engineering~\citep{sahoo2024systematic}. Our prompt optimization was primarily conducted using Gemini 2.5~\citep{deepmind_gemini_v2.5_2025}. First, we directly asked models, including Gemini, Claude~\citep{anthropic_claude37_sonnet_2025}, and GPT~\citep{openai_gpt4_1_api_2025} to read the competition requirements and generate our initial prompts. Based on evaluation results, we used the best-performing model as the base model, while also having Gemini 2.5 analyze the excellent aspects from results generated by underperforming prompts to optimize the base prompt. Additionally, users would independently improve models based on their understanding of the problem, or have models self-improve without relying on external results, followed by resubmission to the leaderboard and repetition of the analysis and merging process.

Our experiments show that this strategy produces clear and original ideas in three technical domains and places first in the PBIG leaderboard's automatic evaluation. Human judges likewise favor our outputs in the NLP and Computer Science tracks, though a gap remains in Materials Chemistry.

\section{Related Works} \label{sec:related-works}
\subsection{Idea Generation}
Idea generation receives widespread attention, particularly in scientific discovery. \citet{wang-etal-2024-scimon} proposed a framework that generates sufficiently innovative ideas by continuously comparing ideas with existing papers. \citet{si2025can} demonstrated that LLMs can generate novel research ideas through large-scale experiments and compared them with human ideas. The results showed that LLM-generated ideas exhibit greater novelty but lack feasibility. \citet{meincke2024using} also explored product idea generation and found that AI-generated ideas yield higher purchase intent but lower novelty and greater similarity. However, for top-ranked ideas, AI demonstrated advantages over human-generated ideas. Overall, the literature indicates that AI-generated ideas possess inherent value and are cost-effective.

\subsection{Patent Processing and Business Application}
Patents constitute a critical resource for business intelligence, enabling the extraction of insights into technological trends and competitive landscapes through patent mining and patent landscaping~\citep{yoon2011identifying,tseng2007text,van2023patent}. The application of artificial intelligence (AI) and natural language processing (NLP) has fundamentally transformed traditionally manual processes, enabling the automation of large-scale semantic analysis of patent documents. These computational approaches transcend the limitations of keyword-based methods and facilitate more sophisticated assessments of novelty and identification of strategic opportunities~\citep{jiang2025natural, shomee2024comprehensive}.

The advent of generative AI and LLMs represents a paradigm shift from analytical to generative capabilities in patent-related tasks. Recent studies have demonstrated that LLMs can effectively generate novel invention concepts and refine existing patent drafts~\citep{jiang2024can, wang2024patentformer, kawano2024claimbrush}. Although current applications predominantly target the generation of technical inventions themselves, they indicate substantial potential for LLMs to facilitate downstream innovation activities, including product ideation and business model development.

\subsection{LLM-as-a-Judge}
Evaluating the creativity of LLMs is a non-trivial task~\citep{si2025can}. To systematically assess LLM performance, researchers developed platforms such as Chatbot Arena~\citep{zheng2023judging}, which ranks models through crowdsourced pairwise comparisons; however, relying on human annotators is costly. Consequently, using powerful LLMs as automated evaluators has become a promising alternative. Yet, studies also show that LLMs exhibit their own inherent biases, such as a preference for their own generated outputs, as well as sensitivity to the position and length of the text they evaluate~\citep{zheng2023judging, DBLP:conf/iclr/YeWHCZMGG0CC025}. When sufficient data are available, a viable approach is to train smaller, specialized LLMs that can match the performance of larger, closed-source models for evaluation tasks~\citep{zhu2025judgelm, chiang2023vicuna}. Given the inherent bias of LLMs, the automated evaluation of these models remains a critical challenge.

\section{Problem Definition} \label{sec:problem-definition}
The PBIG task supplies 150 patents—50 each from Natural Language Processing (NLP), Computer Science (CS), and Materials Chemistry (MC). Each patent appears as a JSON file that lists its title, abstract, claims, description, publication number, and other bibliographic fields, together with the original PDF and figure images. Participants must propose one product per patent that exploits the disclosed technology and can plausibly reach the market within three years.

The required submission is a JSON object with four text fields: a product title of at most 100 characters, a product description of at most 300 characters, an implementation outline of at most 300 characters, and a differentiation statement of at most 300 characters. External resources, such as additional patents or web data, may be consulted when generating ideas.

Systems are compared pairwise and ranked with an Elo scheme~\citep{elo1967proposed}. Both LLMs and human experts score each pair on six criteria: \textit{technical validity}, \textit{innovativeness}, \textit{specificity}, \textit{need validity}, \textit{market size}, and \textit{competitive advantage}.

\section{Methodology} \label{sec:methodology}
We adopt a lightweight pipeline that relies solely on the supplied patent text and the generative capacity of LLMs, without external training data or manual feature engineering. Our workflow consists of model selection, prompt construction, length control, minimal domain adaptation, and an internal Elo-style evaluation loop.

\subsection{Base Model Selection}
We compared GPT-4.1~\citep{openai_gpt4_1_api_2025}, GPT-4o~\citep{openai_gpt4o_2024}, Claude 3.7 Sonnet~\citep{anthropic_claude37_sonnet_2025}, and Gemini 2\&2.5~\citep{deepmind_gemini_v2.5_2025}. Taking into account our budget, usage habits, and performance, we chose GPT-4.1 to generate the final results. Different prompts were crafted for each model, and they were not shared across team members at the start of development. GPT-4.1's selection was also influenced by its better performance with the specific prompts developed for it. Overall, this model may not be the best-performing one. Due to the time constraints and conditions of the competition, we did not conduct a more detailed analysis. Although newer reasoning-oriented models such as OpenAI o1~\citep{openai2024openai} and DeepSeek R1~\citep{deepseek-ai2025deepseekr1} have demonstrated strong performance in complex tasks, we did not select them due to their relatively high inference cost and slower response times. Since the PBIG task required evaluating and refining outputs across multiple prompts and settings, low-latency generation was prioritized to enable efficient iteration.

\subsection{Prompt Generation}
We initially generated candidate prompts using different LLMs, guided by the official PBIG guidelines. Instead of directly merging prompts written by team members, we adopted an LLM-assisted refinement strategy. Gemini was instructed to analyze outputs from underperforming prompts, identify effective components, and integrate them into the current best-performing prompt. This process was repeated iteratively to improve prompt quality. Although the loop still involves manual steps, we believe that it can be further optimized and fully automated in the future through systematic prompt exploration and evaluation. 

\subsection{Length Control}
We found that longer system prompts made it harder to constrain output length, even with explicit character limits. Attempts to shorten outputs by post-editing often reduced scores and performed worse than simply truncating the original text. Our final solution was to restate the character limit at the end of the user prompt. This strategy proved effective, possibly because constraints placed closer to the generation starting point are given higher priority.

\subsection{Domain Adaptation}
The prompt tuned on NLP patents served as a base for all domains. For CS and MC, we asked Gemini 2.5 to inject domain-specific terminology into the same prompt. GPT-4.1 then produced the final outputs without additional fine-tuning.

\subsection{Evaluation}
We created an internal leaderboard that mirrors the official Elo scheme~\citep{chatbotarena}.
We implemented only LLM-based evaluation and conducted pairwise comparisons for all six evaluation criteria in a single step, rather than performing separate pairwise comparisons for each criterion.
This approach improved evaluation efficiency and allowed for rapid comparison of multiple prompts.
When comparing two generated outputs, we truncated them according to the required output constraints described in Section \ref{sec:problem-definition}.
To mitigate potential position bias in the comparison, we swapped the positions of the outputs in 50\% of the cases. This can avoid some bias and does not affect computation time. For leaderboard evaluation, we selected Qwen3-8B~\citep{qwen3technicalreport} due to its relatively low cost yet high correlation with GPT-4.1.
To further save evaluation time, we first compared the new results with the previous best results using GPT-4.1-mini and only submitted the results to the leaderboard when improvements were confirmed.

The final submission set was determined based on the final leaderboard evaluation. In the final evaluation, the best-performing models varied across the three domains.

Therefore, we selected a model that achieved a balance between ranking and the degree of length-limit violation. Figure~\ref{fig:prompt_overview} summarises the final prompt structure. The complete prompt is reproduced in Appendix~\ref{sec:final_prompts}.
\begin{figure}[!t]
  \centering
  \begin{AIboxmain}{Prompt overview (excerpt)}
  \small
  \textbf{Role} “You are an expert business strategist and product-innovation analyst …”\\
  \textbf{Mission} Craft exactly one product idea that critically leverages the patent’s core NLP innovation.\\
  \textbf{Evaluation targets}\\
  1) Technical validity 2) Innovativeness 3) Specificity \\
  4) Need validity 5) Market potential 6) Competitive advantage \\
  \textbf{Output format (char limits)}\\
  \verb|"title"| (100) \verb|"product_description"| (300) \\
  \verb|"implementation"| (300) \verb|"differentiation"| (300) \\
  \textbf{Critical constraints}\\
  -- Patent must be indispensable \\
  -- Launch $\leq$ 3 years \\
  -- Strict character limits \\
  -- One idea only \\
  -- Self-check: “Could the value exist without this patent?”
  \end{AIboxmain}
  \caption{Condensed view of the final prompt.  The full two-page version appears in Appendix~\ref{sec:final_prompts}.}
  \label{fig:prompt_overview}
\end{figure}

\section{Results} \label{sec:results}
\subsection{Overall Performance}
Table~\ref{tab:mk2-results} summarizes the evaluation scores of our system (MK2) across the three domains under automatic and human evaluation settings. MK2 consistently performed well across most domains and criteria, except for human evaluation in MC. In the AI automated evaluation component, MK2 demonstrated significant advantages.

\begin{table*}[!t]
  \centering
  \caption{Evaluation scores of MK2 across domains and evaluation types. Boldface marks the best score in each criterion and evaluation type. NLP: Natural Language Processing, CS: Computer Science, MC: Materials Chemistry, Tech Valid: Technical Validity, Spec: Specificity, Need Valid: Need Validity, Innov: Innovativeness, Comp Adv: Competitive Advantage.}
  \label{tab:mk2-results}
  \begin{tabular}{llrrrrrr}
    \toprule
    \textbf{Domain} & \textbf{Evaluation} & \textbf{Tech Valid} & \textbf{Spec} & \textbf{Need Valid} & \textbf{Market Size} & \textbf{Innov} & \textbf{Comp Adv} \\
    \midrule
    \multirow{2}{*}{NLP}
    & Auto  & \textbf{1093} & \textbf{1215} & \textbf{1076} & 1008 & \textbf{1215} & \textbf{1150} \\
    & Human & \textbf{1025} & \textbf{1044} & \textbf{1009} & 921 & \textbf{1103} & \textbf{1035} \\
    \multirow{2}{*}{CS}
    & Auto  & \textbf{1107} & \textbf{1170} & \textbf{1053} & \textbf{1056} & \textbf{1169} & \textbf{1124} \\
    & Human & \textbf{1018} & 995 & \textbf{1074} & 999 & \textbf{1036} & \textbf{1017} \\
    \multirow{2}{*}{MC}
    & Auto  & \textbf{1132} & \textbf{1184} & 1125 & \textbf{1118} & \textbf{1207} & \textbf{1146} \\
    & Human & 1017 & 1010 & 989 & 1013 & 990 & 991 \\
    \bottomrule
  \end{tabular}
\end{table*}

\subsection{Domain-wise Analysis}
In the \textbf{NLP domain}, MK2 obtained the highest scores in five out of six criteria, except for \textit{market size}, under both automatic and human evaluations. The system particularly excelled in \textit{specificity} and \textit{innovativeness}. This result shows that MK2 can produce clear and original ideas grounded in relevant knowledge. The relatively lower score in \textit{market size} suggests room to clarify economic feasibility by adding concrete use cases or specific target segments.

In the \textbf{CS domain}, MK2 obtained the top automatic evaluation scores across all six criteria. High scores in \textit{specificity} and \textit{innovativeness} reflect the system's ability to produce well-grounded and original ideas. Regarding human evaluations, MK2 ranked top among the four criteria. Lower ranks in \textit{specificity} and \textit{market size} suggest a need for better descriptions of technical depth and economic relevance.

In the \textbf{MC domain}, MK2 obtained the highest automatic evaluation scores in five out of six criteria. Strong performance in \textit{specificity} and \textit{innovativeness} indicates that the system can propose technically detailed and novel ideas rooted in scientific content. However, MK2 did not achieve the top score in any criterion under human evaluation. This contrast implies that automatic metrics may not fully capture the scientific rigor expected by domain experts.

\section{Discussion} \label{sec:discussion}
The evaluation results reveal both the strengths and the limitations of MK2. Automatic scores placed MK2 at the top in every domain, and human judges confirmed this superiority in the NLP and CS tasks. These outcomes show that MK2 can produce clear, original ideas that draw on relevant domain knowledge. In contrast, the MC task exposed a gap: MK2 earned high automatic scores yet failed to lead in any criterion under human evaluation. The outputs, although well structured, did not fully satisfy expert expectations for technical accuracy, clarity, or scientific plausibility. This finding underscores the need for stronger domain constraints and validation steps when addressing specialized fields. Considering that our method involves adapting from NLP to other domains, this discrepancy may not stem from our lack of knowledge in MC, but rather from differences in LLMs' understanding across domains.

Item-level inspection added another layer of insight. Human scores fluctuated widely, with some ideas rated highly and others judged poor. Such variability points to the difficulty of consistent expert assessment and highlights the need for more reliable protocols, particularly in technical domains.

These observations raise broader questions about evaluation design. Automatic metrics scale well and often align with human views on creativity, yet they can miss critical aspects of scientific rigor. Human review captures those nuances, but it suffers from subjectivity and inconsistency, especially when feasibility must be judged. A hybrid approach that combines automatic screening with focused expert review may offer a better balance.

In summary, MK2 generates innovative, well-specified ideas in several scientific fields, but refinement is necessary to meet expert standards in highly technical domains. Future work should deepen domain adaptation in generation and develop evaluation frameworks that assess technical feasibility and clarity more reliably.

\section*{Acknowledgments}
This work was partly supported by JST ACT-X Grant Number JPMJAX22A4 and JST Moonshot R\&D Program Grant Number JPMJMS2236.


\appendix
\section{Final Prompts} \label{sec:final_prompts}
The final prompt used in our submission is reproduced in Figure~\ref{fig:final_prompt}.
\begin{figure*}[!t]
  \centering
    \begin{AIbox}{Final prompt (page 1/2)}
    {\scriptsize
    You are an expert business strategist and product-innovation analyst with deep expertise in **NLP technology commercialization**, AI-driven startup ideation, and market analysis. Your mission is to help a world-class AI research team win an international innovation competition by crafting an **outstanding product idea** derived from an NLP-related patent.\\
    You will be provided with a patent in JSON format, containing fields such as 'title', 'abstract', 'claims', and 'description'. Your primary task is to deeply analyze this patent. Produce **exactly ONE** commercially viable, highly specific, and technically feasible product concept that **critically leverages** the patent's core innovation, making it the **irreplaceable foundation** of the product.\\
    \# Your product idea must EXCEL in:\\
    1.  **Technical Validity \& Feasibility** – Launchable within ~3 years; core functionality MUST depend on the identified patented NLP mechanism. *The implementation should clearly justify this 3-year timeline.*\\
    2.  **Innovativeness** – Clearly show how the patent unlocks a new, superior solution. Specify if the innovation lies in solving an existing problem in a *radically more effective/efficient way* OR if it enables a *completely new application/market* previously unfeasible. *This novelty should be a clear thread through your description and differentiation.*\\
    3.  **Specificity** – Pin-pointed target users/industry, precise pain points (ideally quantified, e.g., 'X hours wasted weekly per user', 'Y\% error rate leading to \$Z losses'), and a concrete use case. Avoid overly broad markets (e.g., 'all businesses') or vague pain points (e.g., 'improving efficiency'). Focus on a defined user segment (e.g., 'compliance officers in Tier-2 investment banks') and a specific, compelling problem (e.g., 'reducing false positives in AML transaction monitoring by X\%').\\
    4.  **Need Validity** – Address a compelling, validated pain point. Persuasively explain why the target users will adopt this solution over existing alternatives or inaction, emphasizing the quantifiable benefits.\\
    5.  **Market Potential** – The product should target a sizable, growing, or high-value niche market. This aspect, potentially with an indicative market size (e.g., '\$X billion market'), *or the scale/frequency of the problem*, should be briefly and credibly touched upon in the product description.\\
    6.  **Competitive Advantage** – Demonstrate a durable, significant edge **directly and uniquely enabled by the identified core patented NLP mechanism**, making it difficult for competitors to replicate (e.g., due to patent protection or the unique technical barrier). This advantage must be critical to the target users/industry.\\
    \# Ideation Process:\\
    1.  **Deep Patent Analysis** – **Critically, identify and articulate the single most unique, pivotal, and non-obvious technical mechanism, algorithm, or architectural innovation** detailed within the patent (often found in the 'claims' or 'detailed description' sections). *Consider what makes this specific mechanism distinct from general NLP techniques.* This specific element is the *cornerstone*.\\
    2.  **Market \& Need Identification** – Locate significant unmet needs or "white-space" opportunities where this specific patented breakthrough can deliver outsized, clearly demonstrable (and ideally quantifiable) value. *Consider current market trends and if the patent addresses an accelerating need.* Analyze current solution gaps and their measurable impact.\\
    3.  **Product Conceptualization** – Design a focused product where the **identified patented mechanism** is absolutely indispensable to delivering the core user value and solving the pinpointed, quantified problem.\\
    4.  **Strategic Pitch Formulation** – Craft a concise, compelling, VC-ready pitch for your product idea, ensuring all evaluation criteria **are evident in the output fields below** and met within the specified character limits, emphasizing quantifiable achievements. *Remember, evaluators will assess each criterion, so ensure your output text for each field strongly reflects the relevant criteria.*\\
    \textit{(continue to next page)}
    }
    \end{AIbox}
  \caption{Final prompt used in our submission.} \label{fig:final_prompt}
\end{figure*}

\begin{figure*}[!t]
  \centering
    \begin{AIbox}{Final prompt (page 2/2)}
    {\scriptsize
    \# Output Format – return **only** the JSON object:\\
    \{\\
    "title": "Concise, specific, and catchy product name that hints at its value or core NLP technology (max 100 characters). **Strict adherence to this character limit is mandatory.**",\\
    "product\_description": "Start by clearly stating the target users/industry and their primary, validated pain point, **ideally quantifying it (e.g., 'X hours wasted weekly', 'Y\% inefficiency costing \$Z')**. Then, describe your product as the distinct solution, detailing its key NLP-driven features (explicitly linking them to the patent's innovation, *showcasing its novelty*) and core benefits, **ideally quantifying these benefits (e.g., 'reduces A by B\%', 'saves C hours per user', 'improves accuracy by D\%')**. If credible and concise, briefly weave in an indicative market size (e.g., '\$X billion addressable market') or scale of the problem to underscore the opportunity (max 300 characters). **Strict adherence to this character limit is mandatory.**",\\
    "implementation": "Explain precisely how the *identified core patented NLP mechanism* (from your Deep Patent Analysis, referencing specific aspects or relevant **Claim numbers/details if they directly support the core mechanism** and are illustrative and concise) is integrated as the central component of the product. Detail why it's feasible to commercialize and launch within ~3 years, mentioning necessary supporting AI/ML infrastructure, **key complementary technologies, API integrations, or critical partnerships** if relevant for successful deployment and operation. *Briefly touch upon the development stages or milestones that make the 3-year timeline realistic.* (max 300 characters). **Strict adherence to this character limit is mandatory.**",\\
    "differentiation": "Clearly articulate why this product is uniquely superior to current alternatives (name common categories or specific well-known competitors if applicable). Crucially, highlight how the *identified core patented NLP mechanism* directly delivers a lasting, non-replicable (potentially due to the patent protection itself or the unique technical barrier it creates, e.g., 'competitors cannot legally replicate this specific mechanism'), and **decisive** competitive advantage. Quantify this advantage where possible (e.g., 'delivers X\% better results than competitor Y', 'reduces costs by Z compared to existing methods'). Explain *why this specific advantage is critical* and highly valued by the target users/industry. **If applicable, clearly state how the product addresses a 'white space' or underserved niche in the market, *further emphasizing its innovativeness and unique value proposition* .** (max 300 characters). **Strict adherence to this character limit is mandatory.**"\\
    \}\\
    \# Critical Constraints \& Mindset:\\
    -   **Authentic \& Deep Innovation** – Do NOT merely paraphrase patent text. Synthesize, extend, and innovate *from* the patent's core. The patent's mechanism must be the *enabling technology*, not just an incidental feature.
    -   **Creative but Grounded Application:** While innovative, ensure the product application is a practical and plausible use of the patent's core technology, avoiding overly futuristic or speculative concepts not achievable within the timeframe.\\
    -   **Single Best Idea Focus** – Output exactly **one** product idea in the valid JSON format. No extra text or explanations outside the JSON structure.\\
    -   **Strict Character Limits:** Adhere strictly to all specified character limits for each field. Overages will be truncated by the evaluation system.\\
    -   **Assumed Resources** – Assume standard modern AI/ML tooling, cloud platforms, and the possibility of acquiring relevant (public or licensable) datasets are available if realistic for a ~3-year launch.\\
    -   **Founder's Mentality \& Investor Appeal** – Pitch your idea with the conviction and clarity of a startup founder seeking investment from discerning investors and expert judges. Use strong, persuasive language and emphasize clear, quantifiable value. Be bold yet grounded in technical and market reality.\\
    -   **Patent Indispensability Test (Self-Correction):** Before finalizing, critically assess: 'Could this product's core value proposition and its unique competitive advantages (especially the quantified ones) be achieved effectively *without this specific patent's core mechanism*?' If the answer is 'yes' or 'mostly,' the idea needs refinement until the patent's role is truly indispensable and central to the claimed unique value. *The `implementation` and `differentiation` sections must strongly convey this indispensability.*
    }
    \end{AIbox}
  \caption*{{Figure~\ref{fig:final_prompt}: Final prompt used in our submission (continued).}}
\end{figure*}

\section{Representative Output Samples and Human Scores} \label{sec:appendix-sample-outputs}

\subsection*{NLP Domain}
\begin{itemize}
  \item \textbf{Title:} VerticalIQ: Domain-Adaptive Chatbot for Enterprise IT Helpdesks with Dynamic Confidence Routing
  \item \textbf{Product Description:} Target: Enterprise IT helpdesks (10k+ employees). Pain: 40\% of tickets misrouted, causing avg. 6hr delays/ticket (\$15M/yr loss). Solution: Chatbot uses patented vertical/confidence routing to auto-classify \& resolve queries, reducing misrouting by 70\%, saving \$10M/yr. ITSM market: \$10B+.
  \item \textbf{Implementation:} Integrates patented multi-vertical/confidence mechanism (Claims 1,5,8): user input is routed by adaptive keyword sets per IT domain (e.g., networking, software, hardware), switching verticals in real time. Milestones: 1) Data ingestion \& vertical setup; 2) Confidence model tuning; 3) ITSM integratio
  \item \textbf{Differentiation:} Unlike generic chatbots or static intent models (e.g., ServiceNow Virtual Agent), only VerticalIQ uses patented dynamic vertical/confidence routing, reducing misclassification by 70\%. Patent protection blocks replication, critical for large orgs needing accurate, adaptive IT query resolution.
  \item \textbf{Human Evaluation Scores:}
    \begin{itemize}
      \item Specificity: [3, 3, 4, 4]
      \item Technical Validity: [-, -, 2, 3]
      \item Innovativeness: [-, -, 1, 2]
      \item Competitive Advantage: [-, -, 4, 1]
      \item Need Validity: [1, 5, -, -]
      \item Market Size: [2, 4, -, -]
    \end{itemize}
\end{itemize}

\subsection*{Computer Science Domain}
\begin{itemize}
  \item \textbf{Title:} PrivataQuery: High-Performance Secure SQL Analytics for Multi-Party Financial Data Collaboration
  \item \textbf{Product Description:} For financial consortia needing joint analytics on confidential datasets, PrivataQuery enables secure, efficient SQL queries across encrypted databases. By extracting only valid rows post-operation (per patent), it cuts secure query compute by up to 99.99\%. \$2B+ secure analytics market.
  \item \textbf{Implementation:} Core patented row reduction protocol (Claims 1, 3) is embedded in the secure SQL engine, minimizing dummy row overhead. Built on modern MPC libraries, cloud orchestration, and secure APIs. 3-year plan: prototype, pilot with banks, full SaaS launch. Requires secure infra partners.
  \item \textbf{Differentiation:} Unlike generic MPC DBs (e.g., Sharemind, CypherDB), PrivataQuery's patented valid row extraction slashes compute and latency by orders of magnitude‚Äîenabling practical, scalable secure analytics. Competitors can't match this efficiency due to patent-protected architecture.
  \item \textbf{Human Evaluation Scores:}
    \begin{itemize}
      \item Specificity: [3, 4, 3, 3]
      \item Technical Validity: [2, 3, -, -]
      \item Innovativeness: [3, 2, -, -]
      \item Competitive Advantage: [2, 4, -, -]
      \item Need Validity: [-, -, 5, 3]
      \item Market Size: [-, -, 2, 2]
    \end{itemize}
\end{itemize}

\subsection*{Materials Chemistry Domain}
\begin{itemize}
  \item \textbf{Title:} GearXcelTM: Ultra-Durable, Low-Friction Polyacetal Composite Gears for Automotive Powertrains
  \item \textbf{Product Description:} Automotive Tier 1 suppliers face gear failures from creep/wear (current POM gears: <1000 hr creep rupture, >0.35 friction coeff.). GearXcelTM gears use patented block-copolymer POM + acid-modified glass fiber, achieving >2000 hr creep life, <0.18 friction, 140+ MPa strength. \$2B+ global market.
  \item \textbf{Implementation:} Utilizes claim 2/3: ABA block-copolymer POM, acid-modified glass fiber, and surface-enriched low-MW PE. Commercialization leverages existing twin-screw extrusion/injection molding lines; 3-year launch feasible via pilot runs, ISO/automotive validation, and OEM co-development.
  \item \textbf{Differentiation:} Conventional POM gears lack >2.90 (œÉ-65)/GF\% ratio, <0.2Œºm resin coating, or surface PE enrichment‚Äîleading to lower durability and higher wear. Patent-protected interface engineering yields >100\% creep life and 40\% lower friction, enabling downsizing and warranty cost reduction for OEMs.
  \item \textbf{Human Evaluation Scores:}
    \begin{itemize}
      \item Specificity: [4, 4, 3, 4]
      \item Technical Validity: [3, -, 2, 1]
      \item Innovativeness: [1, 1, 1, -]
      \item Competitive Advantage: [2, 1, 1, -]
      \item Need Validity: [2, 1, 2, 3]
      \item Market Size: [2, 1, 2, 5]
    \end{itemize}
\end{itemize}

\section{Participants}
Other participants in this shared task include \citeauthor{yoshiyasu2025nsnlp}, \citeauthor{kanumolu2025agentideate}, \citeauthor{terao2025collaborative}, \citeauthor{hoshino2025multiagent}, and \citeauthor{shimanuki2025selfimprovement}~\cite{yoshiyasu2025nsnlp,kanumolu2025agentideate,terao2025collaborative,hoshino2025multiagent,shimanuki2025selfimprovement}. We thank all participants for their valuable contributions to this workshop.
\end{document}